\documentclass[a4paper]{llncs} 
\pdfoutput=1 

\usepackage[T1]{fontenc}
\usepackage[utf8]{inputenc}
\usepackage{lmodern}

\usepackage{amsmath,amssymb}

\usepackage{graphicx}
\usepackage{color}
\usepackage{subcaption}
\usepackage{algorithm}
\usepackage{algorithmic}

\usepackage[english]{babel}
\usepackage{booktabs}
\usepackage{xspace}
\usepackage{complexity} 

\usepackage{hyperref}

\newcommand{\pdftitle}{A Bayesian Network Model \\ for Interesting Itemsets}

\hypersetup{
  pdftitle={\pdftitle},
  pdfauthor={},
  plainpages=false,
  colorlinks=true,
  linkcolor=blue,
  citecolor=blue,
  urlcolor=blue,
}
\addto\extrasenglish{}
\addto\extrasenglish{}
\addto\extrasenglish{}

\allowdisplaybreaks[3] 

\newcommand{\cf}{\hbox{{cf.}}\xspace}

\newcommand{\eg}{\hbox{{e.g.}}\xspace}
\newcommand{\ie}{\hbox{{i.e.}}\xspace, } 
\newcommand{\st}{\hbox{{s.t.}}\xspace}

\renewcommand{\R}{\mathbb{R}}

\newcommand{\abs}[1]{\lvert #1 \rvert}
\newcommand{\norm}[1]{\lVert #1 \rVert}
\newcommand{\deq}{\mathrel{\mathop:}=}

\renewcommand{\S}{§} 

\newlength{\emstr}
\newcommand{\boldpara}[1]{%
\smallskip%
\par\noindent\textbf{\textit{#1}}\hspace{\emstr}
}%

\usepackage[misc,geometry]{ifsym}

\begin{document} 

\title{\pdftitle}

\author{%
Jaroslav Fowkes (\Letter)
\and 
Charles Sutton
}
\institute{School of Informatics,
University of Edinburgh, Edinburgh, EH8 9AB, UK\\
\texttt{\{jfowkes,csutton\}@inf.ed.ac.uk}}

\maketitle

\begin{abstract}
Mining itemsets that are the most interesting under a statistical model of the underlying data is a commonly used and well-studied technique for exploratory data analysis, with the most recent interestingness models exhibiting state of the art performance. 
Continuing this highly promising line of work, we propose the first, to the best of our knowledge, generative model over itemsets, in the form of a Bayesian network, and an associated novel measure of interestingness.
Our model is able to efficiently infer interesting itemsets directly from the transaction database using structural EM, in which the E-step employs the greedy approximation to weighted set cover. Our approach is theoretically simple,
straightforward to implement, trivially parallelizable and retrieves itemsets whose quality is comparable to, if not better than, existing state of the art algorithms as we demonstrate on several real-world datasets.

\end{abstract} 

\section{Introduction}\label{sec:intro}
Itemset mining is one of the most important problems in data mining,
with applications including market basket analysis,
mining data streams and mining bugs in source code
\cite{aggarwal2014frequent}.
Early work on itemset mining focused on algorithms that identify all itemsets which meet 
a given criterion for pattern quality,
such as all \emph{frequent itemsets} whose support is above a user-specified threshold.
Although appealing algorithmically, the list of frequent itemsets suffers from \emph{pattern explosion}, \ie is
typically long, highly redundant and difficult to understand \cite{aggarwal2014frequent}.
%
In an attempt to address this problem, more recent work focuses on mining \emph{interesting itemsets},
smaller sets of high-quality, non-redundant itemsets that can be examined by a data analyst
to get an overview of the data. Several different approaches have been proposed for this
problem. Some of the most successful recent approaches, such as MTV \cite{mampaey2011tell}, KRIMP \cite{vreeken2011krimp} and SLIM \cite{smets2012slim} are based on the
\emph{minimum description length} (MDL) principle, meaning that they define an encoding scheme for compressing the database
based on a set of itemsets, and search for the itemsets that best compress the data. These methods have been shown to lead to much
less redundant pattern sets than frequent itemset mining.

In this paper, we introduce an alternative, but closely related, viewpoint on interesting itemset mining methods,
by starting with a probabilistic model of the data rather than a compression scheme.
We define a \emph{generative model} of the data, that is, a probability distribution over the 
database, in the form of a Bayesian network model, based on the interesting itemsets. 
To infer the interesting items, we use a probabilistic learning approach 
that directly infers the itemsets that best explain the underlying data. 
Our method, which we call the \emph{Interesting Itemset Miner} (IIM)\footnote{\url{https://github.com/mast-group/itemset-mining}},
is to the best of our knowledge, the first {generative model} for interesting itemset mining.

Interestingly, our viewpoint has a close connection 
to MDL-based approaches for mining
 itemsets that best compress the data (Section~\ref{sec:correspondence}). Every probability distribution implicitly defines an optimal compression algorithm, 
and conversely every compression scheme implicitly corresponds to a probabilistic model.
Explicitly taking the probabilistic modelling perspective rather than an MDL perspective has two advantages.
First, focusing on the probability distribution relieves us from specifying the many book-keeping
details required by a lossless code.
Second, the probabilistic modelling perspective allows us to exploit powerful
methods for probabilistic inference, learning, and optimization, such as 
submodular optimization and structural expectation maximization (EM).

The collection of interesting itemsets under IIM can be inferred efficiently using
a structural EM framework \cite{friedman1998bayesian}.
One can think of our model as a probabilistic relative of some of the early work on itemset mining that formulates the task of finding interesting patterns as a covering problem  \cite{geerts2004tiling,vreeken2011krimp},
except that in our work, the set cover problem 
is used to identify itemsets that cover a transaction
\emph{with maximum probability}. The set cover
problem arises naturally within the E step of the EM algorithm.
On real-world datasets we find that the interesting itemsets 
seem to capture meaningful domain structure,
\eg representing phrases such as \emph{anomaly detection} in a corpus of research papers,
or regions such as \emph{western US states} in geographical data.
Notably, we find that IIM returns a much more diverse list of itemsets
than current state of the art algorithms (Table~\ref{tab:redundancy}),
which seem to be of similar quality.
Overall,
our results suggest that the interesting itemsets found by IIM are suitable for manual examination during exploratory data analysis.


\section{Related Work}\label{sec:lit}
Itemset mining was first introduced by Agrawal and Srikant \cite{agrawal1994fast}, along with the Apriori algorithm, in the context of market basket analysis which led to a number of other algorithms for frequent itemset mining including Eclat and FPGrowth. Frequent itemset mining suffers from \emph{pattern explosion}: a huge number of highly redundant frequent itemsets are retrieved if the given minimum support threshold is 
too low. One way to address this is to mine \emph{compact representations} of frequent itemsets such as maximal frequent, closed frequent and non-derivable itemsets with efficient algorithms such as 
CHARM \cite{zaki2002charm}. However, even mining such compact representations does not fully resolve the problem of pattern explosion (see Chapter 2 of \cite{aggarwal2014frequent} for a survey of frequent itemset mining algorithms).

An orthogonal research direction has been to mine \emph{tiles} instead of itemsets, 
\ie subsets of rows \emph{and columns} of the database viewed as binary transaction by item matrices. The analogous approach is then to mine \emph{large tiles}, \ie sub-matrices with only 1s whose area is greater than a given 
minimum area threshold. The Tiling algorithm \cite{geerts2004tiling} is an example 
of an efficient implementation that uses the greedy algorithm for set cover.
Note that there is a correspondence between tiles and itemsets: every large tile is a closed 
frequent itemset and thus algorithms for large tile mining also suffer from pattern explosion to some extent.

In an attempt to tackle this problem, modern approaches to 
itemset mining have used the \emph{minimum description length} (MDL) principle to find the set of itemsets that best summarize the database.
MTV \cite{mampaey2012summarizing} uses MDL coupled with a \emph{maximum entropy} (MaxEnt) model to mine the most informative itemsets. 
MTV mines the set of top itemsets with the highest likelihood under the model 
via an efficient convex bound that allows many candidate itemsets to be pruned and employs a method for more efficiently inferring the model itself. 
Due to the partitioning constraints necessary to keep computation feasible, MTV typically only finds in the order of tens of itemsets, whereas IIM has no such restriction.

KRIMP \cite{vreeken2011krimp} employs MDL to find the subset of frequent itemsets that yields the best lossless compression of the database. While in principle this could be formulated as a set cover problem, the authors employ a fast heuristic that does not allow the itemsets to overlap (unlike IIM) even though one might expect that doing so could lead to better compression. 
In contrast, IIM employs a set cover framework to identify
a set of itemsets that cover a transaction with highest probability.
The main drawback of KRIMP is the need to mine a set of frequent itemsets in the first instance, which is addressed by the SLIM algorithm \cite{smets2012slim}, an extension of KRIMP that mines itemsets directly from the database, iteratively joining co-occurring itemsets such that compression is maximised.

The MaxEnt model can also be extended to tiles, here known as the \emph{Rasch} model, 
and, unlike in the itemset case, inference takes polynomial time.
Kontonasios and De Bie \cite{kontonasios2010information} use the Rasch model to find the most surprising 
set of \emph{noisy tiles} (\ie sub-matrices with predominantly 1s but some 0s) by computing the likelihood of tile entries covered by the set. The 
inference problem then takes the form of weighted budgeted maximum set cover, which 
can again be efficiently solved using the greedy algorithm. The problem of Boolean matrix factorization can be viewed as finding  a set of frequent noisy tiles which form a low-rank approximation to the data \cite{miettinen2008discrete}.   

The MINI algorithm \cite{gallo2007mini} finds the itemsets with the highest surprisal under statistical independence models of items and transactions from a precomputed set of closed frequent itemsets. OPUS Miner \cite{webb2014efficient} is a branch and bound algorithm for mining the top \emph{self-sufficient} itemsets, \ie those whose frequency cannot be explained solely by the frequency of either their subsets or of their supersets.  

In contrast to previous work, IIM maintains a generative model, in the form of a 
Bayesian network, \emph{directly} over itemsets as opposed to indirectly over items. 
Existing Bayesian network models for itemset mining 
\cite{he2012bayesian,jaroszewicz2004interestingness} have had limited success as modelling dependencies between the items makes inference for larger datasets prohibitive. 
In IIM inference takes the form of a weighted set 
cover problem, which can be solved efficiently using the greedy algorithm 
(Section~\ref{sec:inference}).

The structure of IIM's statistical model is similar to existing
models in the literature such as Rephil (\cite{murphy2012machine}, \S
26.5.4) for topic modelling and QMR-DT \cite{shwe1991probabilistic} for medical diagnosis.
Rephil is a multi-level graphical model 
used in Google's AdSense system.
QMR-DT is a bi-partite graphical model
used for inferring significant diseases based on medical findings.
However, the main contribution of our paper is to show that a binary latent
variable model can be useful for selecting itemsets for exploratory data
analysis.

\section{Interesting Itemset Mining}\label{sec:mining}
In this section we will formulate the problem of identifying a set of 
interesting itemsets that are useful for explaining a database of transactions. 
First we will define some preliminary concepts and notation. An \emph{item} $i$ 
is an element of the universe $U = \{1,2,\dotsc,n\}$ that indexes database
attributes. A \emph{transaction} $X$ is a subset of the universe $U$ and an
\emph{itemset} $S$ is simply a set of items $i$. The set of
interesting itemsets $\mathcal{I}$ we wish to determine is therefore a subset of
the power set (set of all possible subsets) of the universe. Further, we say
that an itemset $S$ is \emph{supported} by a transaction $X$ if $S \subseteq X$.

\subsection{Problem Formulation}

Our aim in this work is to infer a set of interesting itemsets $\mathcal{I}$ from a database of transactions.
By \emph{interesting},
we mean a set of itemsets that will best help a human analyst to understand the important properties of the database,
that is, interesting itemsets should reflect the important probabilistic dependencies among items, while being sufficiently concise and non-redundant
that they can be examined manually.
These criteria are inherently qualitative, reflecting the fact that the goal of data mining is to build human
insight and understanding.
In this work, we formalize  
{interestingness} as those itemsets that best explain the transaction database under a \emph{statistical model} of itemsets. Specifically we will use a \emph{generative} model, \ie a model that starts with a set of interesting itemsets $\mathcal{I}$ and from this set generates the transaction database. Our goal is then to infer the most likely generating set $\mathcal{I}$ under our chosen generative model. We want the model to be as simple as possible yet powerful enough to capture correlations between transaction items. A simple such model is to iteratively sample itemsets $S$ from $\mathcal{I}$ and let their union form a transaction $X$. Sampling $S$ from $\mathcal{I}$ uniformly would be uninformative, but if we associate each interesting itemset $S \in \mathcal{I}$ with a probability $\pi_S$, we can sample the indicator variable $z_S \sim $ Bernoulli$(\pi_S)$ and include $S$ in $X$ if $z_S = 1$. We formally define this generative model next.

\subsection{Bayesian Network Model}\label{sec:model}
We propose a simple directed graphical model for generating a database of transactions $X^{(1)},\dotsc,X^{(m)}$ from a set $\mathcal{I}$ of interesting
itemsets. 
The parameters of our model are Bernoulli
probabilities $\pi_S$ for each interesting itemset $S \in \mathcal{I}$. The
generative story for our model is, independently for each transaction $X$:
\begin{enumerate}
 \item For each itemset $S \in \mathcal{I}$, decide independently whether to 
include $S$ in the transaction, \ie sample 
 \[ z_S \sim \text{ Bernoulli}(\pi_S). \]
 \item Set the transaction to be the set of items in all the itemsets selected 
 above: 
 \[ X = \bigcup_{S | z_S = 1} S.\]
\end{enumerate}
Note that the model allows individual items to be generated multiple 
times from different itemsets, \eg \textit{eggs} could be generated both as
part of a breakfast itemset \{\textit{bacon, eggs}\} and as as part of a cake
itemset \{\textit{flour, sugar, eggs}\}.

Now given a set of itemsets $\mathcal{I}$, let $\mathbf{z}, \boldsymbol\pi$ denote 
the vectors of $z_S,\pi_S$ for all $S \in \mathcal{I}$. Assuming 
$\mathbf{z}, \boldsymbol\pi$ are fully determined, it is evident from the 
generative model that the probability of generating a transaction $X$ is%
\begin{equation}\label{eq:model}
 p(X, \mathbf{z}|\boldsymbol\pi) =
\begin{cases}
\,\prod_{S \in \mathcal{I}}
\pi_S^{z_S}(1-\pi_S)^{1-z_S} \! &\text{if $X = \bigcup_{z_S = 1} S$,} \\
\,0 \! &\text{otherwise}
\end{cases}.
\end{equation}


\subsection{Inference}\label{sec:inference}
Assuming the parameters $\boldsymbol\pi$ in the model are known, we can infer 
$\mathbf{z}$ for a specific transaction $X$ by maximizing the posterior
distribution
$p(\mathbf{z}|X,\boldsymbol\pi)$ over $\mathbf{z}$:
\begin{equation}\label{eq:ml}
\max_{\mathbf{z}} \prod_{S \in \mathcal{I}} \pi_S^{z_S}(1-\pi_S)^{1-z_S} \qquad\st \; X = \bigcup_{S | z_S = 1} S.
\end{equation}
Taking logs and rewriting \eqref{eq:ml} in a more standard form we obtain
\begin{equation}\label{eq:ilp}
\begin{split}
\min_{\mathbf{z}} &\sum_{S \in \mathcal{I}} z_S \left(-\ln(\pi_S)\right) + (1-z_S)\left(-\ln(1 - \pi_S)\right) \\
\:\st \; &\sum_{S|i \in S}z_S \ge 1 \quad\, \forall \, i \in X, \quad\; z_S \in \{0,1\} \quad \forall \, S \in \mathcal{I}
\end{split}
\end{equation}
which is (up to a penalty term) the weighted set-cover problem (see \eg \cite{korte2012combinatorial}, \S
16.1) with
weights $w_S \in \mathbb{R}^+$ given by $w_S \deq -\ln(\pi_S)$.
This is an NP-hard problem in general and so impractical to solve 
directly in practice. It is important to note that the weighted set cover
problem is a special case 
of minimizing a linear function subject to a submodular constraint,\footnote{Note
that the posterior $p(z|X)$ would not be submodular if we were to use a noisy-OR
model for the conditional probabilities.}
which we formulate as follows
(\cf \cite{young2008greedy}). Given the set of interesting itemsets
$\mathcal{T} \deq \{S \in \mathcal{I}\,|\,S \subseteq X\}$
that support the transaction,
a real-valued weight $w_S$ for each itemset $S \in \mathcal{T}$ and a 
non-decreasing submodular function $f:2^\mathcal{T} \to \R$, the aim is to find 
a covering $C \subset \mathcal{T}$ of minimum total weight, \ie such that 
$f(\mathcal{C}) = f(\mathcal{T})$ and $\sum_{S \in \mathcal{C}}w_S$ is 
minimized. For weighted set cover we simply define $f(\mathcal{C})$ to be the
number of items in $\mathcal{C}$, \ie
$f(\mathcal{C}) \deq \abs{\cup_{S \in \mathcal{C}} S}.$
Note that $f(\mathcal{T}) = \abs{X}$ by construction.

We can then approximately solve the weighted set cover problem
\eqref{eq:ilp} using the greedy approximation algorithm for submodular 
functions. 
The greedy algorithm builds a covering $\mathcal{C}$ by repeatedly 
choosing an itemset $S$ that minimizes the weight $w_S$ divided by the number
of items in $S$ not yet covered by the covering. In order to minimize CPU time
spent solving the weighted set cover problem, we cache the
itemsets and coverings for each transaction as needed.

It has been shown \cite{chvatal1979greedy} that the greedy algorithm achieves
a $\ln\abs{X} +1$ approximation ratio to the weighted set cover problem and 
moreover the following inapproximability theorem shows that this ratio is essentially 
the best possible.
\begin{theorem}[Feige \cite{feige1998threshold}]
There is no $(1-o(1))\ln\abs{X}$-approximation algorithm to the weighted set 
cover problem unless $\NP \subseteq \DTIME(\abs{X}^{O(\log\log\abs{X})})$, \ie 
unless $\NP$ has slightly superpolynomial time algorithms.
\end{theorem} 
The runtime complexity of the greedy algorithm 
is $O(\abs{X}\abs{\mathcal{T}})$, however by maintaining a priority queue this can 
be improved to $O(\abs{X}\log\abs{\mathcal{T}})$ (see
\eg \cite{cormen2001introduction}).
Note that there is also an $O(\abs{X}\abs{\mathcal{T}})$-runtime primal-dual 
approximation algorithm \cite{bar1981linear}, however this has an approximation
order of 
$f = \max_{i}\abs{\{S\,|\,i \in S\}}$, \ie the frequency of the most frequent 
element, which would be worse in our case.

\subsection{Learning}
Given a set of itemsets $\mathcal{I}$, consider now the case where both
variables $\mathbf{z}, \boldsymbol\pi$ in the model are unknown. In this case
we can use the hard EM algorithm
\cite{dempster1977maximum} for parameter estimation with latent variables. The
hard EM algorithm in our case is merely a simple layer on top of the inference
algorithm \eqref{eq:ilp}. Suppose there are $m$ transactions
$X^{(1)},\dotsc,X^{(m)}$
with supporting sets of itemsets $\mathcal{T}^{(1)},\dotsc,\mathcal{T}^{(m)}$,
then the hard EM algorithm is given in Algorithm~\ref{alg:em}.
To initialize $\boldsymbol\pi$, a natural choice is simply the support (\ie relative
frequency) of each itemset in $\mathcal{I}$.

\begin{algorithm}[tb]
\caption{\textsc{Hard-EM}}\label{alg:em}
\algsetup{indent=1em}
\begin{algorithmic}
\REQUIRE Set of itemsets $\mathcal{I}$ and initial probability estimates
$\boldsymbol\pi^{(0)}$
\STATE $k \leftarrow 0$
\REPEAT
\STATE $k \leftarrow k+1$
\STATE \textsc{E-step: } $\forall \, X^{(j)}$ solve \eqref{eq:ilp} 
to get $z_S^{(j)} \;\, \forall\, S \in
\mathcal{T}_j$
\STATE \textsc{M-step: } $\pi_S^{(k)} \leftarrow \frac{1}{m} \sum_{j=1}^m
z_S^{(j)} \quad \forall\, S \in \mathcal{I}$
\UNTIL{$\norm{\boldsymbol\pi^{(k-1)} - \boldsymbol\pi^{(k)}} > \varepsilon\,$}
\STATE Remove from $\mathcal{I}$ itemsets $S$ with $\pi_S = 0$
\RETURN $\mathcal{I}, \boldsymbol\pi^{(k)}$
\end{algorithmic}
\end{algorithm}

\subsection{Inferring new itemsets}
We infer new itemsets using structural EM \cite{friedman1998bayesian}, \ie
we add a candidate itemset $S'$ to $\mathcal{I}$ if doing so 
improves the optimal value $\overline{p}$ of the problem \eqref{eq:ilp} averaged
across transactions. 
Interestingly, there is an implicit regularization
effect here. Observe from \eqref{eq:ilp} that when a new candidate $S'$ is
added to the model, a corresponding term $\ln (1-\pi_{S'})$ is added to the
log-likelihood of all transactions that $S'$ does not support. For large databases,
this amounts to a significant penalty on candidates. 

To get an estimate of maximum benefit to including
candidate $S'$, we must carefully choose an initial value of
$\pi_{S'}$ that is not too low, to avoid getting stuck in a local optimum.
To infer a good $\pi_{S'}$, we force the candidate $S'$ to explain all
transactions it supports by initializing $\pi_{S'} = 1$ and update $\pi_{S'}$
with the probability corresponding to its actual usage once we have inferred
all the coverings. Given a set of itemsets $\mathcal{I}$ and corresponding
probabilities
$\boldsymbol\pi$ along with transactions $X^{(1)},\dotsc,X^{(m)}$,
each iteration of the 
structural EM algorithm is is given in Algorithm~\ref{alg:sem} above.
\begin{algorithm}[tb]
\caption{\textsc{Structural-EM} (one iteration) 
}\label{alg:sem}
\algsetup{indent=1em}
\begin{algorithmic}
\REQUIRE Itemsets $\mathcal{I}$, probabilities $\boldsymbol\pi$,
optima $p^{(j)}$ of \eqref{eq:ilp} $\forall \, X^{(j)}$
\STATE Set profit  $\overline{p} \leftarrow \frac{1}{m}
\sum_{j=1}^m p^{(j)}$
\REPEAT 
\STATE Generate candidate $S'$ using \textsc{Candidate-Gen}
\STATE $\mathcal{I} \leftarrow \mathcal{I}\cup
\{S'\}, \, \pi_{S'} \leftarrow 1$
\STATE \textsc{E-step: } $\forall \, X^{(j)}$ solve \eqref{eq:ilp} to get $z_S^{(j)} \;\, \forall\, S \in \mathcal{T}_j$
\STATE \textsc{M-step: } $\pi_S' \leftarrow \frac{1}{m} \sum_{j=1}^m
z_S^{(j)} \quad \forall\, S \in \mathcal{I}$
\STATE $\forall \, X^{(j)}$, solve \eqref{eq:ilp} using $\pi_S',z_S^{(j)} \;\, \forall\, S \in \mathcal{T}_j$ to get the optimum $p^{(j)}$
\STATE Set new profit $\overline{p}' \leftarrow \frac{1}{m} \sum_{j=1}^m
p^{(j)}$
\STATE $\mathcal{I} \leftarrow \mathcal{I}\setminus \{S'\}$
\UNTIL{$\overline{p}' \le \overline{p}$} \COMMENT{until one good candidate
found} 
\STATE $\mathcal{I} \leftarrow \mathcal{I}\cup \{S'\}$
\RETURN $\mathcal{I}, \boldsymbol\pi'$
\end{algorithmic}
\end{algorithm}

In practice, we cache the set of candidates that have been rejected by the 
\textsc{Structural-EM} function to avoid reconsidering them. 

\subsection{Candidate generation}\label{sec:candgen}
The \textsc{Structural-EM} algorithm (Algorithm~\ref{alg:sem}) requires a method to generate 
new candidate itemsets $S'$ that are to be considered for inclusion in the set 
of interesting itemsets $\mathcal{I}$. One possibility would be to use the 
Apriori algorithm to recursively suggest larger itemsets starting from singletons, 
however preliminary experiments found this was not the most efficient method. 
For this reason we take a slightly different approach and recursively
combine the interesting itemsets in $\mathcal{I}$ with the \emph{highest
support first} (Algorithm~\ref{alg:combine}). In this way our candidate generation
algorithm is more likely to propose viable candidate itemsets earlier and in
practice we find that this heuristic works well. 
We did try pruning potential itemset pairs to join using a $\chi^2$-test, however 
this substantially slowed down the algorithm and barely improved the model likelihood.

\begin{algorithm}[tb]
\caption{\textsc{Candidate-Gen}}\label{alg:combine}
\algsetup{indent=1em}
\begin{algorithmic}
\REQUIRE Itemsets $\mathcal{I}$, cached supports $\boldsymbol\sigma$, queue length $q$
\IF{$\nexists$ priority queue $\mathcal{Q}$ for $\mathcal{I}$}
\STATE Initialize $\boldsymbol\sigma$-ordered priority queue $\mathcal{Q}$
\STATE Sort $\mathcal{I}$ by decreasing itemset support using $\boldsymbol\sigma$
\FOR{all distinct pairs $S_1,S_2 \in \mathcal{I}$, highest
ranked first}
\STATE Generate candidate $S' = S_1\cup S_2$
\STATE Cache support of $S'$ in $\boldsymbol\sigma$ and add $S'$ to $\mathcal{Q}$
\STATE \textbf{if} $\abs{\mathcal{Q}}=q$ \textbf{break} 
\ENDFOR
\ENDIF
\STATE Pull highest-ranked candidate $S'$ from $\mathcal{Q}$
\RETURN $S'$
\end{algorithmic}
\end{algorithm}

In order to determine the supports of the
itemsets to be combined, we store the transaction database in
a Memory-Efficient Itemset Tree (\textsc{MEI-Tree}) \cite{fournier2013meit} and
query the tree for the support of a given itemset. A \textsc{MEI-Tree} stores itemsets in
a tree structure according to their prefixes in a memory efficient manner.
To minimize the
memory usage of the \textsc{MEI-Tree} further, we first sort the items in
order of decreasing support (as in the FPGrowth algorithm) as this
often results in a sparser tree \cite{han2000mining}. Note that a \textsc{MEI-Tree} is
essentially an FP-tree \cite{han2000mining} with node-compression and without
node-links for nodes containing the same item.
An itemset support query on the \textsc{MEI-Tree} efficiently searches the tree for
all occurrences of the given itemset and adds up their supports (see
Figure 4 in \cite{fournier2013meit} for the actual algorithm).
With the wide availability of 100GB+ shared memory systems, it is reasonable to expect the \textsc{MEI-Tree} to fit into memory for all but the largest of datasets. The queue length parameter in the \textsc{Candidate-Gen} algorithm effectively imposes a limit on the number of iterations the algorithm can spend suggesting candidate itemsets. 


\begin{algorithm}[tb]
\caption{\textsc{IIM} (Interesting Itemset Miner)}\label{alg:iim}
\algsetup{indent=1em}
\begin{algorithmic}
\REQUIRE Database of transactions $X^{(1)},\dotsc,X^{(m)}$
\STATE Initialize $\mathcal{I}$ with singletons, $\boldsymbol\pi$ with their supports
\STATE Build \textsc{MEI-Tree} from transaction database
\WHILE{not converged}
\STATE Add itemsets to $\mathcal{I}, \boldsymbol\pi$ using \textsc{Structural-EM}
\STATE Optimize parameters for $\mathcal{I}, \boldsymbol\pi$ using \textsc{Hard-EM}
\ENDWHILE
\RETURN $\mathcal{I},\boldsymbol\pi$
\end{algorithmic}
\end{algorithm}

\subsection{Mining Interesting Itemsets}
Our complete interesting itemset mining (IIM) algorithm is given in Algorithm~\ref{alg:iim}.
Note that the \textsc{Hard-EM} parameter optimization step need not be
performed at every iteration, in fact it is more efficient to
suggest several candidate itemsets before optimizing the parameters. As all operations on transactions in our algorithm are
trivially parallelizable, we perform the $E$ and $M$-steps in both
the hard and structural EM algorithms in parallel.

\subsection{Interestingness Measure}\label{sec:interestingness}
Now that we have inferred the model variables $\mathbf{z},\boldsymbol\pi$, we are able to use them to rank the retrieved itemsets in $\mathcal{I}$. There are two natural rankings one can employ, and both have their strengths and weaknesses. The obvious approach is to rank each itemset $S \in \mathcal{I}$ according to its probability under the model $\pi_S$, however this has the disadvantage of strongly favouring frequent itemsets over rare ones, an issue we would like to avoid. Instead, we prefer to rank the retrieved itemsets according to their \emph{interestingness} under the model, that is the ratio of transactions they explain to transactions they support. One can think of interestingness as a measure of how necessary the itemset is to the model: the higher the interestingness, the more supported transactions
the itemset explains. Thus interestingness provides a more balanced measure than probability, at the expense of missing some frequent itemsets that only explain some of the transactions they support. We define interestingness formally as follows.
\begin{definition}
The \emph{interestingness} of an itemset $S\in \mathcal{I}$ retrieved by IIM
(Algorithm~\ref{alg:iim}) is defined as
\[
 int(S) = \frac{\sum_{j=1}^m
z_S^{(j)}}{supp(S)}
\]
and ranges from $0$ (least interesting) to $1$ (most interesting).
\end{definition}
Any ties in the ranking can be broken using the itemset probability $\pi_S$.

\subsection{Correspondence to existing models}\label{sec:correspondence}
There is a close connection between probabilistic models and the MDL principle \cite{mackay:book}. Given a probabilistic model $p(X | \boldsymbol\pi, \mathcal{I})$ of a single transaction, by Shannon’s theorem the optimal code for the model will encode $X$ using approximately $-\log_2 p(X | \boldsymbol\pi, \mathcal{I})$ bits. So by finding a set of itemsets that maximizes the probability of the data, we are also finding itemsets that minimize description length. Conversely, any encoding scheme implicitly defines a probabilistic model: given an encoding scheme $E$ that assigns each transaction $X$ to a string of $L(X)$ bits, we can define $p(X | E) \propto 2^{-L(X)}$, and then $E$ is an optimal code for $p(X | E)$. Interpreting previous MDL-based itemset mining methods in terms of their implicit probabilistic models provides interesting insights into these methods.

MTV uses a MaxEnt distribution over itemsets $S \in \mathcal{I}$, which for a transaction $X$ can be written (\cf \cite{mampaey2012summarizing}):
\[
p(X) = \pi_0 \prod_{S \in \mathcal{I}} \pi_S^{\mathbf{1}_X(S)}
\]
where the indicator function $\mathbf{1}_X(S) = 1$ if $X$ supports $S$ and $0$ otherwise. 
Thus if an itemset is present in the MaxEnt model \emph{it must be used} to explain a supported transaction, contrast this with IIM \eqref{eq:model} where there is a latent variable $z_S^{(j)}$ for each transaction $X^{(j)}$ that \emph{infers if an itemset is used} to explain the transaction. 

KRIMP by contrast, uses an itemset independence model, which for an itemset $S \in \mathcal{I}$ is given by (\cf \cite{vreeken2011krimp}):
\[
p(S) = \sum_{j=1}^m z_S^{(j)} \Bigg/ \sum_{I \in \mathcal{I}} \sum_{k=1}^m z_I^{(k)}
\]
where the $z_S^{(j)}$, and therefore itemset coverings for $X^{(j)}$, are determined using a 
\emph{heuristic approximation}. That is, unlike IIM, the itemset coverings are not chosen to maximise the probability under the statistical model. Instead, for each transaction $X$, frequent itemsets $S \in \mathcal{I}$ are chosen in order of \emph{decreasing size and support} and added to the covering if they improve the compression, until all elements of $X$ are covered. Additionally, itemsets in the covering are not allowed to overlap, in contrast to IIM which does allow overlap if it is deemed necessary. 

SLIM uses the same approach as KRIMP but iteratively finds the candidate itemsets $S$ directly from the dataset. It employs a greedy heuristic to do this: starting with a set of singleton itemsets $\mathcal{I}$, pairwise combinations of itemsets in $\mathcal{I}$ are considered as candidate itemsets $S$ in order of highest estimated compression gain. IIM uses a very similar heuristic that iteratively extends itemsets by the most frequent itemset in its candidate generation step (Section~\ref{sec:candgen}). 

However, IIM is different from these methods in that they all contain an explicit penalty term for the description length of the itemset database, which corresponds to a prior distribution $p(\mathcal{I})$ over itemsets. We did not find in practice that an explicit prior distribution was necessary but it would be possible to trivially incorporate it. Also, if we view IIM as an MDL-type method, not only the presence of an itemset, but also its absence is explicitly encoded (in the form of $(1-\pi_S)^{1-z_S^{(j)}}$ in \eqref{eq:model}). As a result, there is an implicit penalty for adding too many patterns to the model and one does not need to use a code table which would serve as an explicit penalty for greater model complexity.

One can also think of IIM as a probabilistic tiling method: each interesting itemset $S \in \mathcal{I}$ can be thought of as a binary submatrix of transactions for which $z_S=1$ by items in $S$, where the choice of items and transactions in the tile are \emph{inferred directly} from IIM's statistical model. That is, IIM formulates the inference problem \eqref{eq:ilp} as a \emph{weighted set cover} for \emph{each transaction} where the weights correspond to \emph{itemset probabilities}.
This is in contrast to existing tiling methods: Geerts et al.\ \cite{geerts2004tiling} find $k$ tiles covering the largest number of database entries and is thus an instance of \emph{maximum coverage}.  Kontonasios and De Bie \cite{kontonasios2010information} extend this to inferring a covering of noisy tiles using \emph{budgeted maximum coverage}, that is, finding a covering that maximizes the sum of the \emph{surprisal} of each tile, under a MaxEnt model constrained by expected row and column margins,
subject to the sum of the \emph{description lengths} of each tile being smaller than a given budget.

\section{Numerical Experiments}\label{sec:numerics}

\begin{figure}[tb]
  \centering
  \begin{minipage}[t]{0.47\textwidth}
  \centering
  \includegraphics[scale=0.28]{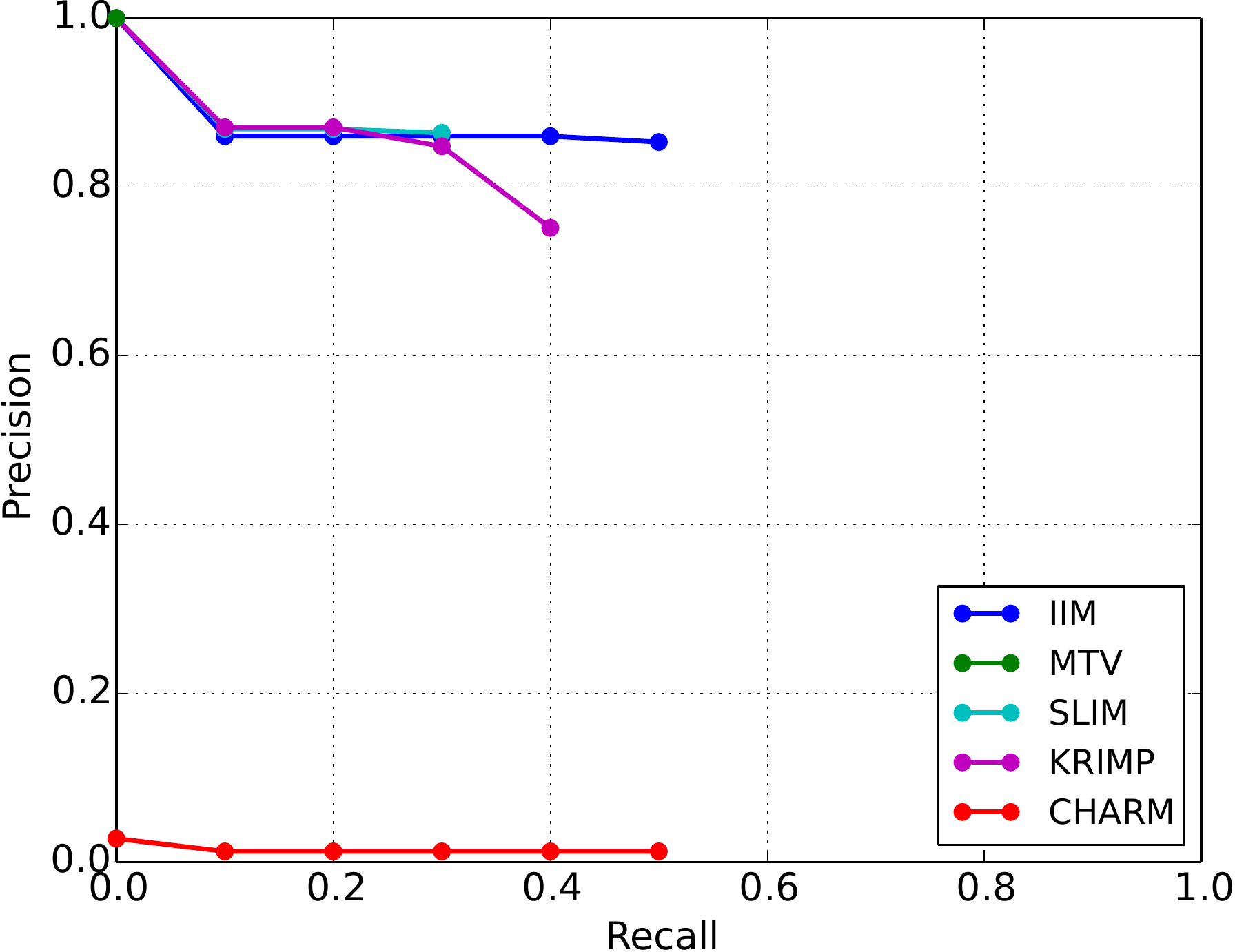}
  \caption{Precision against recall for
each algorithm on our synthetic database, using
the top-$k$ itemsets as a threshold.\protect\footnotemark[3]}
\label{fig:pr}
  \end{minipage}\hfill
  \begin{minipage}[t]{0.47\textwidth}
  \centering
  \includegraphics[scale=0.28]{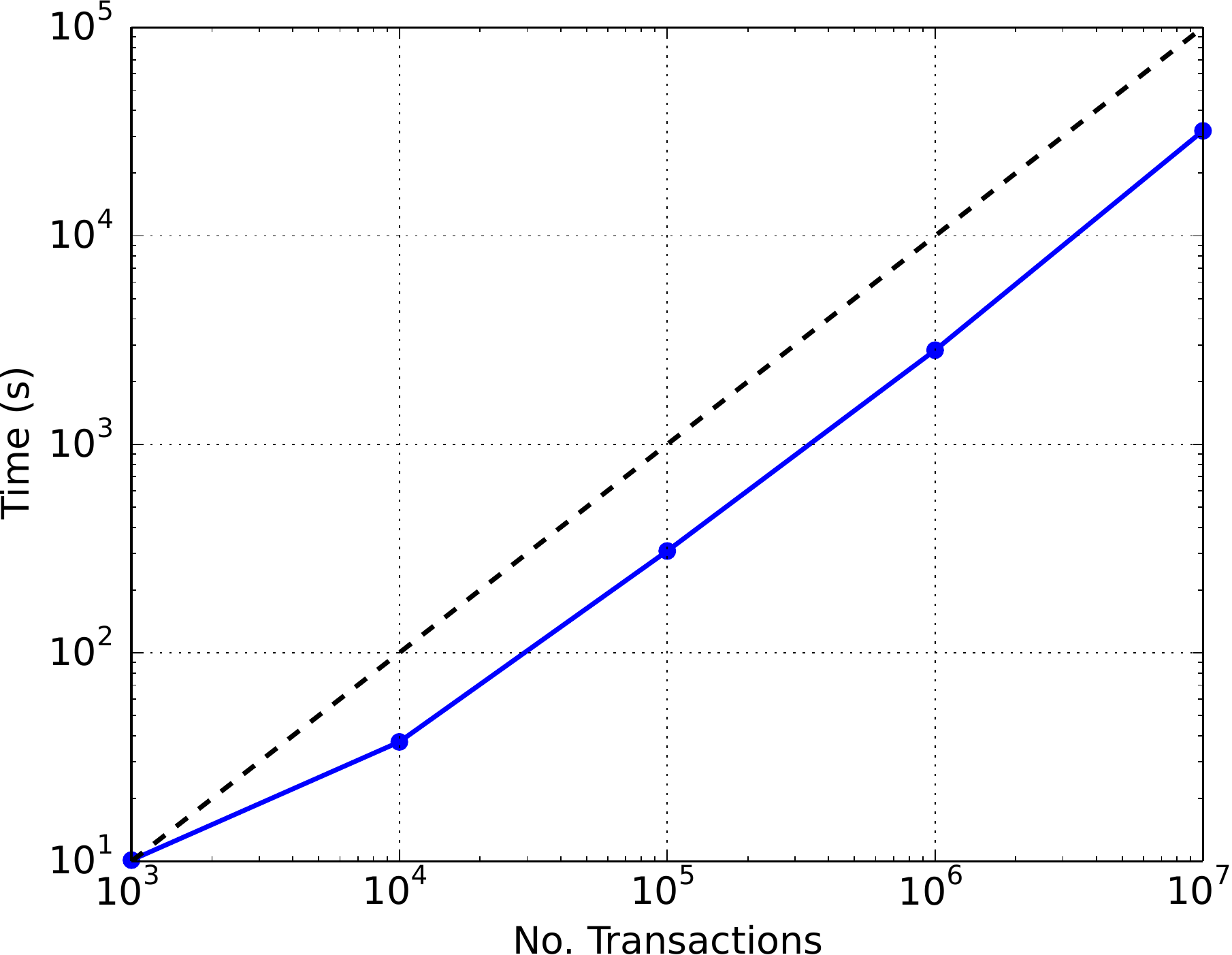}
  \caption{IIM scaling as the number of transactions in our synthetic database
    increases.}
  \label{fig:scaling}
  \end{minipage}
\end{figure}
In this section we perform a comprehensive qualitative and quantitative
evaluation of IIM. On synthetic datasets we show that IIM returns a list of itemsets that 
is largely non-redundant, contains few spurious correlations and scales linearly with the number of transactions. 
On a set of real-world datasets we show that IIM finds itemsets that are much less redundant than
state of the art methods, while being of similar quality.
 
\footnotetext[3]{Each curve is the 11-point
interpolated precision \ie the interpolated precision at 11 equally spaced recall points between $0$ and $1$ (inclusive), see \cite{manning2008introduction}, \S8.4 for details.}

\boldpara{Datasets} We use five real-world datasets in our numerical evaluation (Table~\ref{tab:datasets}). The plants dataset \cite{plants} is a list of plant species and the U.S.\ or Canadian states where they occur. The mammals dataset \cite{mitchell1999atlas} consists of presence records of European mammals in $50 \times 50$ km geographical areas. The retail dataset  consists of anonymized market basket data from a Belgian retail store \cite{goethals2004fimi}.
The ICDM dataset \cite{kontonasios2010information} is a list of ICDM paper abstracts where each item is a stemmed word, excluding stop-words. The Uganda dataset consists of Facebook messages taken from a set of public Uganda-based pages with substantial topical discussion 
over a period of three months. Each transaction in the
dataset is an English language message and each item is a stemmed English word
from the message.

\begin{table}[b]
  \centering
\begin{minipage}[b]{0.49\textwidth}
\caption{Summary of the real datasets used and IIM results after $1,000$
iterations. \dag\,excluding singleton itemsets.}
\footnotesize\centering
  \begin{tabular}{lrr|rr}
  \toprule
  Dataset & Items & Trans.\ \, & \, $\abs{\mathcal{I}}$\dag & Runtime \\
  \midrule
  ICDM & $4,976$ & $859$ \, & $798$ & $163$ min \\
  Mammals & $194$ & $2,670$ \, & $359$ & $22$ min \\
  Plants & $70$ & $34,781$ \, & $259$ & $27$ min \\
  Retail & $16,470$ & $88,162$ \, & $957$ & $941$ min \\
  Uganda & $33,278$ & $124,566$ \, & $928$ & $1086$ min \\
    \bottomrule
  \end{tabular}
\label{tab:datasets}
\end{minipage}\hfill
\begin{minipage}[b]{0.49\textwidth}
\caption{IID for the top $50$ non-singleton itemsets returned by the algorithms. *returned less than $50$ non-singleton itemsets.}
\footnotesize\centering
  \begin{tabular}{lrrrrr}
  \toprule
   & ICDM & Mam.\ & Plant & Retail & Ugan.\ \\
   \cmidrule{2-6}
  \textbf{IIM} & $\mathbf{4.00}$ & $\mathbf{7.42}$ & $\mathbf{4.80}$ & $\mathbf{3.26}$ & $\mathbf{3.78}$ \\
  MTV & $3.14$  & *$5.50$ & *$5.00$ & $2.52$ & *$1.60$ \\
  SLIM & $2.12$ & *$1.76$ & *$1.77$ & $1.44$ & $2.08$ \\
  KRIMP & $2.56$ & $1.94$ & $1.88$ & $1.34$ & $2.26$ \\
  CHARM & $1.42$ & $1.44$ & $1.50$ & $1.32$ & $1.72$ \\
  \bottomrule
  \end{tabular}
\label{tab:redundancy}
\end{minipage}
\end{table}

\boldpara{IIM Results} We ran IIM on each dataset for $1,000$ iterations with a priority queue size of $100,000$ candidates. The runtime and number of non-singleton itemsets returned is given in Table~\ref{tab:datasets} (right). We also investigated the scaling of IIM as the number
of transactions in the database increases, using the model trained on the
plants dataset from Section~\ref{sec:spuriousness} to generate synthetic transaction
databases of various sizes. We then ran IIM for $100$ iterations on these
databases and one can see in Figure~\ref{fig:scaling} that the scaling is
linear as expected. Our prototype implementation can process one million
transactions in 30 seconds on 64 cores each iteration, so there is reason to
hope that a more highly tuned implementation could scale to even larger datasets. All experiments were performed on a machine with 64 AMD Opteron 6376 CPUs and 256GB of RAM.

\boldpara{Evaluation Criteria} We will evaluate IIM along with MTV, SLIM, KRIMP and CHARM with $\chi^2$-test ranking according to the following criteria:
\begin{enumerate}
\item \textit{Spuriousness} -- to assess the degree of spurious correlation in the mined set of itemsets.
\item \textit{Redundancy} -- to measure how redundant the mined set of itemsets is.
\item \textit{Interpretability} 
-- to informally assess how meaningful and relevant the mined itemsets actually are.
\end{enumerate}
Note that we chose not to compare to the tiling methods from \cite{geerts2004tiling,kontonasios2010information}
as they have been shown to underperform on the ICDM dataset \cite{mampaey2012summarizing}.

\subsection{Itemset Spuriousness}\label{sec:spuriousness} 
The set-cover formulation of the IIM algorithm \eqref{eq:ilp} naturally favours
adding itemsets to the model whose items co-occur in the transaction database.
One would therefore expect IIM to largely avoid suggesting itemsets of
uncorrelated items and so generate more meaningful itemsets. To verify
this is the case and validate our inference procedure, we check if IIM is able to recover the itemsets it used generate a synthetic database. To obtain a realistic synthetic database, we sampled $10,000$ transactions from the IIM generative model trained on the plants dataset. We were then able to measure the precision and recall for each algorithm, \ie the fraction of mined itemsets that are generating and the fraction of generating itemsets that are mined, respectively. We used a minimum support of $0.0575$ for all algorithms (except IIM) as used in \cite{mampaey2012summarizing} for the plants dataset. Figure~\ref{fig:pr} shows the precision-recall curve for each algorithm using the top-$k$ mined itemsets (according to each algorithm's ranking) as a threshold.  One can clearly see that IIM was able to mine about $50\%$ of the generating itemsets and almost all the itemsets mined were generating. This not only provides a good validation of IIM’s inference procedure and underlying generative model but also demonstrates that IIM returns few spurious itemsets. For comparison, SLIM and KRIMP exhibited very similar behaviour to IIM whereas MTV returned a very small set of generating itemsets. The set of top itemsets mined by CHARM contained many itemsets that were not generating.
It is not our intention to draw conclusions about the performance of the other algorithms as this experimental setup naturally favours IIM. Instead, we compare the itemsets from IIM with those from MTV, SLIM and KRIMP on real-world data in the next sections.


\subsection{Itemset Redundancy}
We now turn our attention to evaluating whether IIM returns a less redundant
list of itemsets than the other algorithms on real-world datasets.
A suitable measure of redundancy for a single itemset is the minimum symmetric
difference between it and the other itemsets in the list. Averaging this
across all itemsets in the list, we obtain the \emph{average inter-itemset
distance} (IID). We therefore ran all the algorithms on the datasets in Table~\ref{tab:datasets}.
This enabled us to calculate, for each dataset, the IID
of the top $50$ non-singleton itemsets, which we report in
Table~\ref{tab:redundancy}. For CHARM, we took the top $50$ non-singleton 
itemsets ranked according to $\chi^2$ from the top $100,000$ frequent itemsets it returned 
(as the $\chi^2$ calculation would be prohibitively slow otherwise).
One can clearly see that the top IIM itemsets have a
larger IID on average, and
are therefore less redundant, than the KRIMP, SLIM or CHARM itemsets. The 
top CHARM $\chi^2$-ranked itemsets are the most redundant as expected.
On all datasets, the IIM itemsets are less redundant than those mined by
the other methods, with only one exception.
On the Plants dataset, MTV is slightly less redundant than IIM,
but this is because MTV is unable to return $50$ items on this dataset,
instead returning only $21$.

\subsection{Itemset Interpretability}  
For the datasets in Table~\ref{tab:datasets} we can directly interpret the mined itemsets and informally assess how meaningful and relevant they are.  
  
\boldpara{ICDM Dataset}
We compare the top ten non-singleton itemsets mined by the algorithms in Table~\ref{tab:abstracts} (excluding KRIMP whose itemsets are similar for space reasons).
The mined patterns are all very informative, containing technical concepts such as \emph{support vector machine} and common phrases such as \emph{pattern discovery}. The IIM itemsets suggest the stemmer used to process the dataset could be improved, as we retrieve \{\textit{parameter, parameters}\} and \{\textit{sequenc, sequential}\}. 

\begin{table*}[tb]
 \caption{Top ten non-singleton ICDM itemsets as found by IIM, MTV and SLIM.}     
  \footnotesize\centering
  \begin{tabular}{ccc}
  \toprule
  \textrm{IIM} & \textrm{MTV} & \textrm{SLIM} \\
  \cmidrule(r){1-1} \cmidrule{2-2} \cmidrule(l){3-3} 
  associ rule & experiment result & inform model \\
  local global  & synthetic real & cluster algorithm \\
  support vector machin svm  & real datasets &  larg effici \\
  parameter parameters & pattern discov & perform set \\
  anomali detect & associ rule mine & propos problem \\
  sequenc sequential & frequent pattern mine algorithm & method set \\
  linear discriminant analysi & train classifi & associ rule \\
  synthetic real life & address problem & problem result \\
  background knowledg & classifi class & approach base method \\
  semi supervised & machin learn & base method set \\
  \bottomrule
 \end{tabular}
\label{tab:abstracts}
\end{table*}
    
\boldpara{Plants and Mammals Datasets} 
For both datasets, all algorithms find itemsets that are spatially coherent,
but as we showed in Table~\ref{tab:redundancy}, those returned by IIM are far less redundant. 
Our novel interestingness measure enables IIM to rank correlated itemsets above singletons 
and rare itemsets above frequent ones, in contrast to the other algorithms. For example, for the plants dataset, the top itemset retrieved by IIM is \{\textit{Puerto Rico, Virgin Islands}\} whereas MTV returns \{\textit{Puerto Rico}\}, not associating it with the \emph{Virgin Islands} (which are adjacent) until the 20th ranked itemset.
For the mammals dataset, the top two non-singleton IIM itemsets are a group of four mammals that coexist in Scotland and Ireland and a group of ten mammals that coexist on Sweden's border with Norway. By contrast, 
the top four SLIM and KRIMP itemsets list some of the most common mammals in Europe (see the supplementary material for details). 

\begin{figure}[tb]
\centering
  \begin{subfigure}[b]{0.49\textwidth}
  \centering
  \includegraphics[scale=0.325]{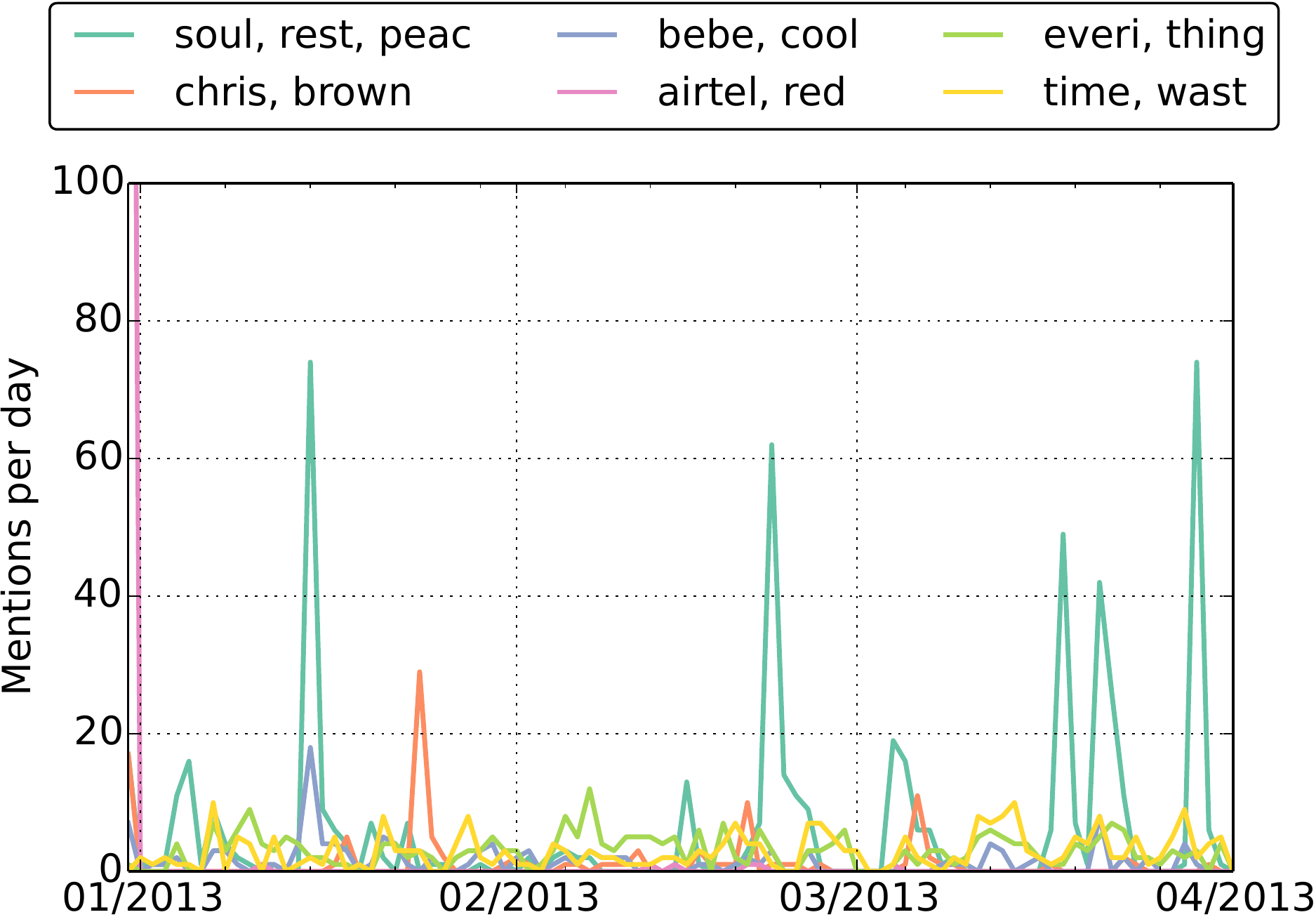}
  \end{subfigure}\hfill
  \begin{subfigure}[b]{0.49\textwidth}
  \centering
  \includegraphics[scale=0.325]{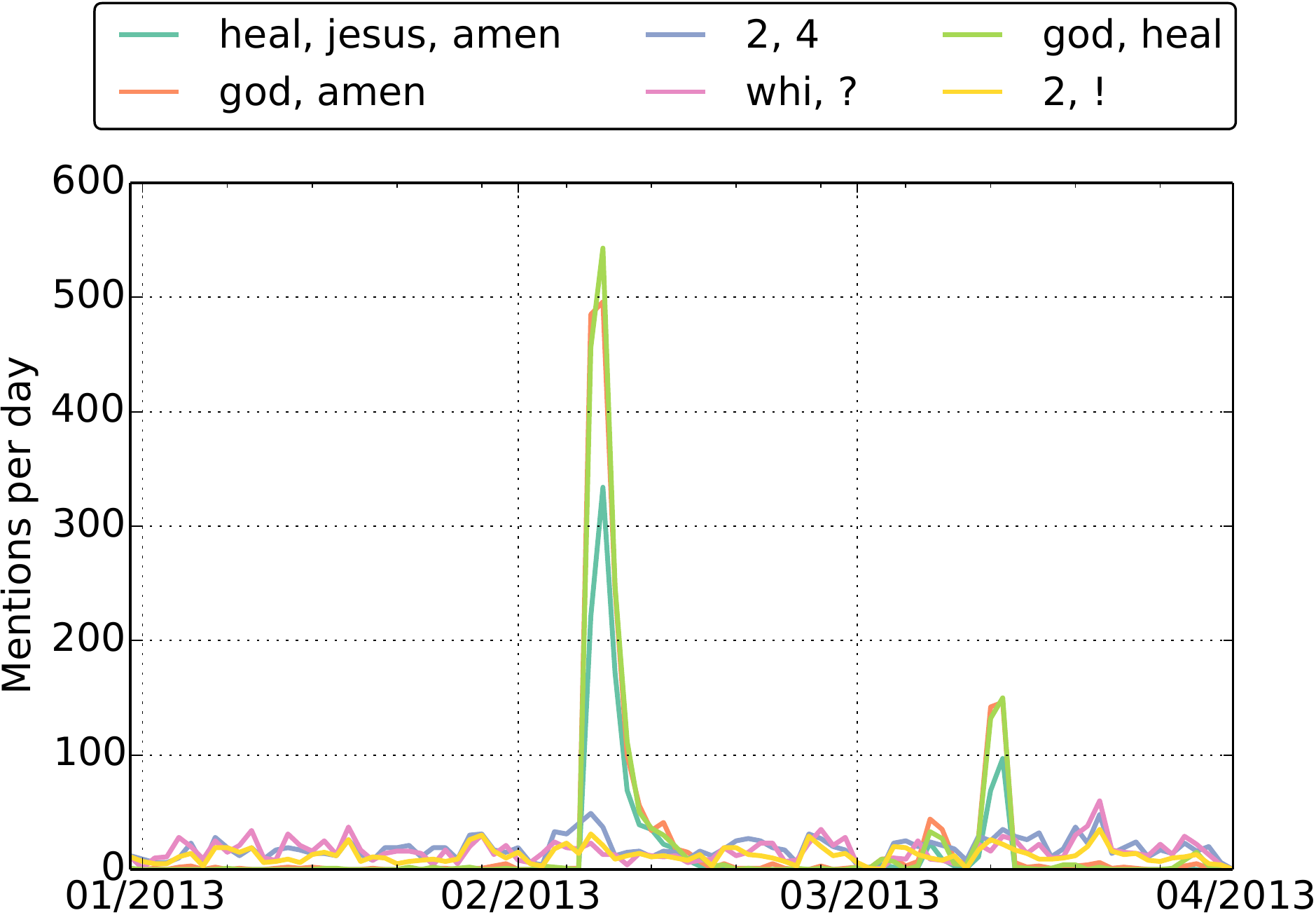} 
  \end{subfigure}
  \caption{Mentions per day of the top six non-singleton IIM (left) and MTV (right) itemsets from the Uganda messages dataset over three months.}
  \label{fig:uganda}
\end{figure}

\boldpara{Uganda Dataset}
The top six non-singleton itemsets found by the algorithms
are shown in Table~\ref{tab:uganda_top}; the IIM itemsets provide much more information
about the topics of the messages than those from the other algorithms.
\begin{table}[tb]
\caption{Top six non-singleton Uganda itemsets for each algorithm.}
 \footnotesize\centering
  \begin{tabular}{cccc}
  \toprule
  IIM & MTV & SLIM & KRIMP \\
  \cmidrule(r){1-1} \cmidrule{2-2} \cmidrule(l){3-3} \cmidrule(l){4-4}
  soul, rest, peace & \; heal, jesus, amen & !, ?  & whi, ? \\
  chris, brown & god, amen & 2, 4 & ?, ! \\
  bebe, cool & 2, 4 & whi, ? & 2, 4  \\
  airtel, red & whi, ? & god, amen & wat, ? \\
  everi, thing & god, heal & da, dat & time, ! \\
  time, wast & 2, ! & heal, jesus, amen & \; soul, rest, peace  \\
    \bottomrule
  \end{tabular}
\label{tab:uganda_top}
\end{table}
Figure~\ref{fig:uganda} (left) plots the
mentions of each of the top IIM itemsets per day.
As one can see, usage of the top itemsets displays temporal structure
(and exhibits spikes of popularity), even though our model does not
explicitly capture this. Of particular
interest are the large spikes of \{\textit{soul, rest, peac}\}
corresponding to notable deaths: wealthy businessman James Mulwana on the 15th
January, President Museveni's father on the 22nd February and
six school students in a traffic accident on the 29th March.
Also of interest are the $285$ mentions of \{\textit{airtel, red}\} on New Year's Eve
corresponding to mobile provider Airtel's Red Christmas competition for 10K worth of airtime. 
The spike of \{\textit{bebe, cool}\} on the 15th January corresponds to the Ugandan musician's 
wedding announcement and the spike on the 24th January of \{\textit{chris, brown}\} refers to many enthusiastic mentions of the
popular American singer that day. The last two itemsets capture common phrases.

In comparison, the top-six MTV itemsets are plotted in Figure~\ref{fig:uganda} (right). One can see that the itemsets \{\textit{heal, jesus, amen}\};\{\textit{god, amen}\} and \{\textit{god, heal}\} substantially overlap and are strongly correlated with each other, sharing a large spike 
on the 8th February and a smaller spike on the 11th March. The remaining itemsets exhibit no interesting spikes as one would expect. 
The top six SLIM and KRIMP itemsets in Table~\ref{tab:uganda_top} all displayed random time evolution, as one would expect, except for the religious ones we have already encountered.

\section{Conclusions}
We presented a generative model that directly infers itemsets that best explain a
transaction database along with a novel model-derived measure of interestingness 
and demonstrated the efficacy of our approach on both
synthetic and real-world databases.
In future we would like to extend our approach to directly inferring the association rules implied by the itemsets and parallelize our approach to large clusters so that we can efficiently scale to much larger databases.

\small 
\subsubsection*{Acknowledgements.}

This work was supported by the Engineering and Physical Sciences Research Council (grant number EP/K024043/1).	
We thank John Quinn for sharing the Uganda data.

\small
\bibliography{pkdd2016}
\bibliographystyle{splncs03}

\end{document}